\documentclass[letterpaper, 10 pt, conference]{ieeeconf}  

\IEEEoverridecommandlockouts                              

\usepackage{mathrsfs}
\usepackage{algorithm}
\usepackage{comment}
\usepackage[noend]{algpseudocode}
\usepackage{amsmath}
\usepackage{amssymb}
\usepackage{array}
\usepackage{booktabs}
\usepackage{caption}
\usepackage{dblfloatfix}
\usepackage{enumerate}
\usepackage{gensymb}
\usepackage{graphicx}
\usepackage{multirow}
\usepackage{makecell}
\usepackage{indentfirst}
\usepackage{tabularx}
\usepackage{threeparttable}
\usepackage[utf8]{inputenc}
\usepackage{subcaption}
\usepackage{mathtools}

\usepackage{color}

\usepackage{tikz}

\newcommand\copyrighttext{%
  \footnotesize \textcopyright 2020 IEEE.  Personal use of this material is permitted.  Permission from IEEE must be obtained for all other uses, in any current or future media, including reprinting/republishing this material for advertising or promotional purposes, creating new collective works, for resale or redistribution to servers or lists, or reuse of any copyrighted component of this work in other works.}
\newcommand\copyrightnotice{%
\begin{tikzpicture}[remember picture,overlay]
\node[anchor=south,yshift=10pt] at (current page.south) {\fbox{\parbox{\dimexpr\textwidth-\fboxsep-\fboxrule\relax}{\copyrighttext}}};
\end{tikzpicture}%
}

\title{\LARGE \bf
Aerial Manipulation using Model Predictive Control \\for Opening a Hinged Door
}

\author{Dongjae Lee$^{1}$, Hoseong Seo$^{1}$, Dabin Kim$^{1}$, and H. Jin Kim$^{2}$
\thanks{*This material is based upon work supported by the Ministry of Trade, Industry \& Energy(MOTIE, Korea) under Industrial Technology Innovation Program. No.10067206, ‘Development of Disaster Response Robot System for Lifesaving and Supporting Fire Fighters at Complex Disaster Environment’.}%
\thanks{$^{1}$Dongjae Lee, Hoseong Seo, and Dabin Kim are graduate students of the Department of Mechanical and Aerospace Engineering, Seoul National University, Seoul, South Korea 
        {\tt\small \{ehdwo713, hosung37, dabin404\}@snu.ac.kr}}%
\thanks{$^{2}$ H. Jin Kim is a faculty of the Department of Mechanical and Aerospace Engineering, Seoul National University, Seoul, South Korea 
        {\tt\small hjinkim@snu.ac.kr}}%
}

\begin{document}

\maketitle
\copyrightnotice
\thispagestyle{empty}
\pagestyle{empty}

\begin{abstract}
Existing studies for environment interaction with an aerial robot have been focused on interaction with static surroundings. However, to fully explore the concept of an aerial manipulation, interaction with moving structures should also be considered. In this paper, a multirotor-based aerial manipulator opening a daily-life moving structure, a hinged door, is presented. In order to address the constrained motion of the structure and to avoid collisions during operation, model predictive control (MPC) is applied to the derived coupled system dynamics between the aerial manipulator and the door involving state constraints. By implementing a constrained version of differential dynamic programming (DDP), MPC can generate position setpoints to the disturbance observer (DOB)-based robust controller in real-time, which is validated by our experimental results.

\end{abstract}

\section{INTRODUCTION}
\label{sec: introduction}
Physical interaction with a surrounding environment has been a research topic of growing interest in aerial robotics. \cite{ruggiero2018aerial} Most studies have been carried out with a hardware platform called unmanned aerial manipulator (UAM), an unmanned aerial vehicle attached with one or more manipulators, with which various real-life applications are investigated such as maintenance tasks \cite{wopereis2018multimodal} and autonomous sensor installation and retrieval operation \cite{hamaza2019sensor}. One compelling employment could be an interaction with a moving structure in the surroundings. Although most of the latest works on UAM physical interaction \cite{alexis2016aerial,wopereis2018multimodal,bodie2019omnidirectional} have focused on an inspection on a static structure, further expansion of the capability of an aerial robot cannot be overlooked. If we can facilitate a UAM to move any movable surrounding structure, a more exhaustive exploration can be performed by accessing once unreachable places, pushing or pulling a movable structure; also, a more active response to disaster or rescue operation can be achieved with its augmented versatility.  

There are at least two additional issues in addressing a movable structure compared to an interaction with a static counterpart: 1) structure dynamics, and  2) collision avoidance. Contrary to a static structure, a movable structure contains its own dynamics which often entails significant force/torque reaction. Without proper modeling or estimation on the movement of the structure, it cannot be guided to the desired position; furthermore, collision avoidance with this dynamic structure cannot be guaranteed which is necessary for safe operation.  

As part of such topics, this paper handles the problem of a UAM opening a hinged door. It deals with a particular application of a multirotor-based UAM operating a movable structure like doors, windows, and heavy cargo. This problem requires additional consideration on the movement of the structure which would involve intrinsic constraints, which is, in our case, a hinge constraint. Moreover, crash prevention with a dynamic structure, the door, and a static structure, a doorframe, is considered along with a self-collision avoidance in generating a collision-free trajectory. After modeling the combined dynamics of the UAM and the hinged door, we applied model predictive control (MPC) to ensure dynamic feasibility and collision avoidance of a generated trajectory. State constraints which are formulated from kinematic relationships are imposed on the optimal control problem, and the generated trajectory from MPC is tracked by a disturbance observer-based robust controller designed in \cite{kim2017robust}.

\begin{figure}
    \centering
    \includegraphics[width=0.7\linewidth]{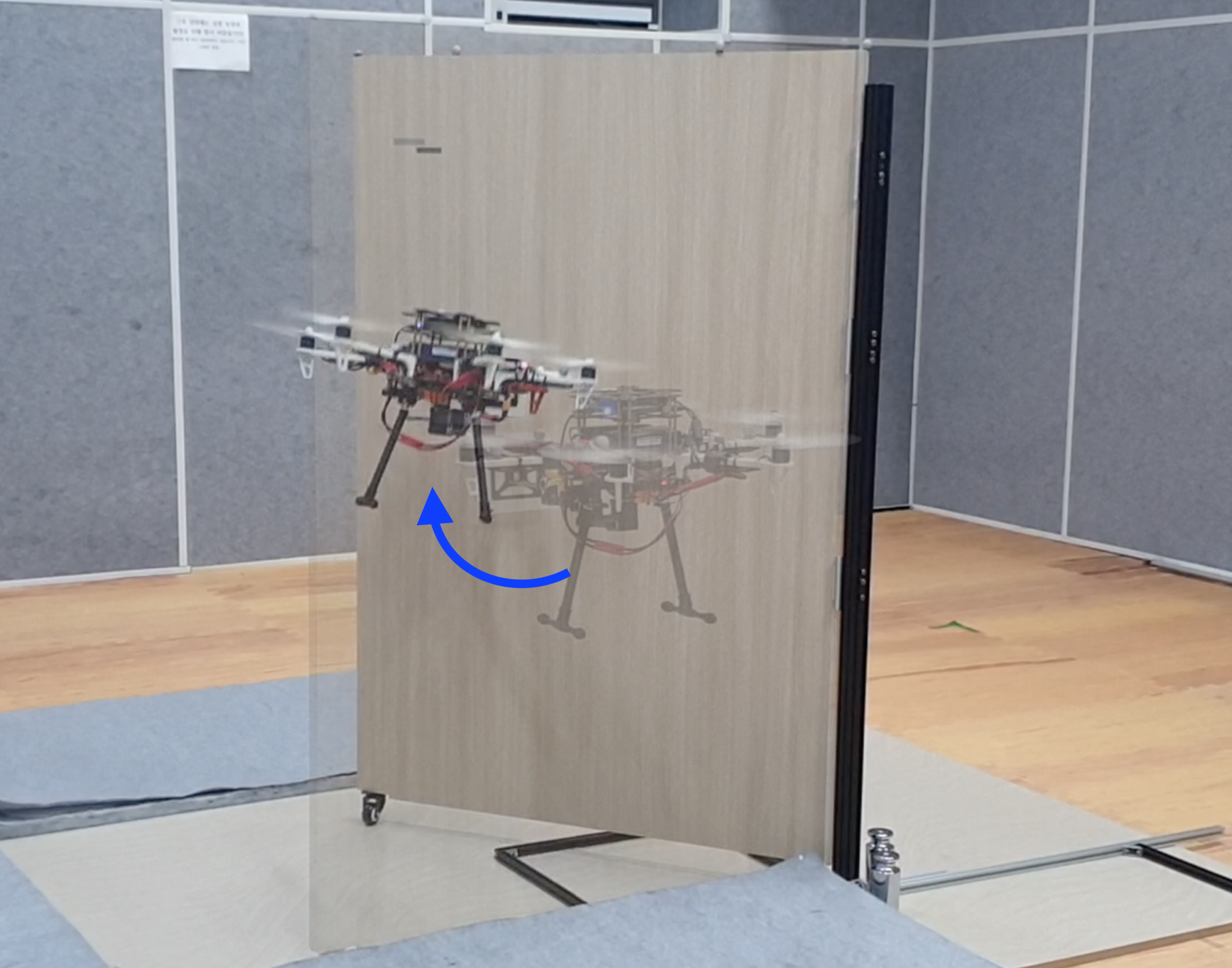}
    \caption{A composite image of an aerial manipulator opening a hinged door in the direction of the blue arrow. A transparent figure is the initial state of the system, and a vivid figure represents the desired state of the system.}
    \label{fig:thumbnail}
    \vspace{-0.3cm}
\end{figure}

\subsection{Related Works}
\label{subsec: related works}
In \cite{wopereis2018multimodal}, tool handling aerial manipulator is suggested while in contact with a vertical surface. For a contact-based inspection task, a motion planning algorithm presented in \cite{tognon2018control} is validated with experiments using a multidirectional thrust aerial vehicle \cite{tognon2019truly}. An omnidirectional aerial vehicle for contact-based inspection on a curved surface is presented in \cite{bodie2019omnidirectional} with state estimation and disturbance rejection. However, none of these considers a moving structure, and since the planner suggested in \cite{tognon2018control} is based on a sampling-based method, it cannot be applied for an online scenario.

There exist some studies about an aerial manipulator coping with a movable structure. In \cite{orsag2017dexterous}, a dual-armed aerial manipulator is employed to perform three different tasks, and the one about a movable structure is a valve-turning task. However, in this scenario, only a rotational motion in the z-axis is required for the task, and certainly, more issues have to be considered for less constrained tasks. In \cite{kim2015operating}, the author presents an experiment of an aerial manipulator operating an unknown drawer. Coupled dynamics between the UAM and the drawer are derived, and the desired force is computed through the velocity of the drawer. However, a structure with only a translational motion is considered, and collision avoidance is not explicitly addressed. In \cite{tsukagoshi2015aerial}, an aerial vehicle opens a hinged door with a proposed mechanism. Though suitable for this particular purpose, versatility that a general UAM contains seems to lack in this approach.          

MPC has been utilized not only as an optimal controller for aerial vehicles but also as an optimal planner. In \cite{darivianakis2014hybrid}, an explicit MPC is used to generate offline control policies for an aerial manipulator interacting with a static structure. Later, in \cite{neunert2016fast}, MPC with SLQ is presented for an online optimal control input computation. Similar techniques are applied in many other papers to produce an online optimal trajectory for a multirotor with a suspended load \cite{son2018model}, and for a multirotor with a network delay \cite{jang2019networked}. In this paper, MPC is utilized as a sub-optimal planner generating a trajectory complying with dynamics and constraints.

\subsection{Contributions}
\label{subsec: contributions}
To the best of our knowledge, this paper presents the first attempt for a UAM to open a hinged door while avoiding collisions. Simplified dynamics and state constraints are formulated to construct an MPC problem, and by adopting a constrained version of differential dynamic programming, dynamically feasible and online-applicable trajectories satisfying constraints are generated. Finally, our proposed approach is validated through real experiments. 

\subsection{Outline}
\label{subsec: outline}
This paper is structured as follows. In Section \ref{sec: problem description}, problem description with general assumptions throughout the paper are explained. Kinematics, dynamics and simplified dynamics are all derived in Section \ref{sec: equations of motion}. In Section \ref{sec: trajectory planning}, problem formulation for MPC with various state constraints guaranteeing collision-free trajectories are presented. Control framework, experimental setup and results are explained in Section \ref{sec: experiments}, while conclusions are drawn in Section \ref{sec: conclusion}. 

\section{PROBLEM DESCRIPTION}
\label{sec: problem description}
This paper considers the problem of a UAM opening a hinged door. Unlike interaction with a static structure, the problem of interacting with a movable structure requires deliberation on the movement of the structure as well. Furthermore, to assure safe operation, constraints on collision avoidance should be reflected. Such collision avoiding constraints in this scenario would involve avoiding collision with a door, a doorframe, and itself. 

To satisfy the need for both being aware of the structure's movement and generating a constraint-abiding trajectory, MPC is applied. An integrated model of the coupled dynamics between the UAM and the hinged door is developed, and three independent constraints are applied to the optimal control problem.

Following assumptions are made throughout this paper:
\begin{itemize}
\item The weight of the robotic arm attached to the bottom of the multirotor is negligible compared to that of the whole system.
\item Servomotors in the robotic arm have negligible dynamics and are assumed to follow the desired velocity command with little time delay.
\item The end-effector's tip and the door's surface are rigidly connected.
\item Physical properties of the door (i.e., mass moment of inertia, width, height, and position) are known.
\label{general assumption}
\end{itemize}
With the followed assumptions, equations of motion and constraints for the system are developed.

\section{EQUATIONS OF MOTION}
\label{sec: equations of motion}
\begin{figure}
    \centering
    \includegraphics[width=0.6\linewidth]{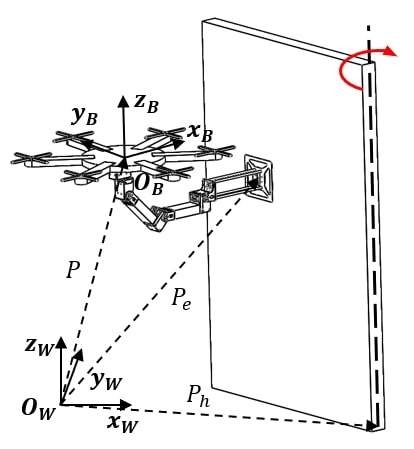}
    \caption{Reference frames of the aerial manipulator in the door opening scenario}
    \label{fig:combined_system}
    \vspace{-0.3cm}
\end{figure}
In this section, equations of motion (EoM) of the whole system including the multirotor-based UAM and the hinged door are derived. A general dynamical model of the system is first introduced by Lagrange equation, and a simplified dynamics is then suggested for real application. 

\subsection{Notations}
\label{subsec: notations}
In order to clarify further derivations, basic notations for variables and parameters that describe the system are predefined. Other additional variables and parameters adopted for notational simplicity are defined across the paper.

The position of the UAM $\mathbf{P} \in {\mathbb{R}}^{3}$, the lower part of the door hinge $\mathbf{P}_h \in {\mathbb{R}}^{3}$, and the end-effector $\mathbf{P}_e \in {\mathbb{R}}^{3}$ are all expressed in the inertial world frame $\mathscr{F}_W$ centered at $O_W$. The attitude of the aerial vehicle and the angle of the door are each denoted by ZYX Euler angle $\mathbf{\Phi} \in {\mathbb{R}}^{3}$ and $\alpha \in {\mathbb{R}}^{1}$ in $\mathscr{F}_W$. For the rotational velocity of the aerial vehicle in the vehicle's body frame $\mathscr{F}_B$ centered at $O_B$, $\mathbf{\Omega} \in {\mathbb{R}}^{3}$ is employed. $O_B$ is positioned at the center of mass (CoM) of the UAM. $\eta_i$, where $i=1,\cdots,4$, describes each servomotor angle composing the multirotor-attached robotic arm in $\mathscr{F}_B$, and they are concatenated as $\mathbf{H}$. Conventionally, $\dot{*}$ denotes an element-wise time derivative of an arbitrary matrix $*$. Among these variables, state variables that describe the coupled EoM of the UAM and the hinged door are defined as $\mathbf{x} = [\mathbf{\Phi} \ \alpha \ \dot{\mathbf{\Phi}} \ \dot{\alpha} \ \mathbf{H}]^T \in {\mathbb{R}}^{12}$, while inputs to the system are defined as $\mathbf{u} = [f_t \ \mathbf{T} \ \mathbf{\dot{H}}_d]^T \in {\mathbb{R}}^{8}$ where $f_t \in {\mathbb{R}}^{1}$, $\mathbf{T} \in {\mathbb{R}}^{3}$, and $\mathbf{\dot{H}}_d \in {\mathbb{R}}^{4}$ each denotes a thrust, torque of the multirotor, and a desired servomotor speed all in $\mathscr{F}_B$. 

Parameters for the UAM are $m_A \in {\mathbb{R}}^{1}, \ I_A \in {\mathbb{R}}^{3 \times 3}, \ l_i \in {\mathbb{R}}^{1}$, which are mass, mass moment of inertia at $O_B$, and $i^{th}$ robotic arm's linkage length, respectively. For the parameters of the door, mass moment of inertia at $\mathbf{P}_h$ $I_D \in {\mathbb{R}}^{1}$, the shortest distance from $\mathbf{P}_e$ to the door hinge $\ D_V \in {\mathbb{R}}^{1}$, the projected distance of the relative position vector between $\mathbf{P}_e$ and $\mathbf{P}_h$ to the door hinge $ \ D_H \in {\mathbb{R}}^{1}$, and the door width $\ D_w \in {\mathbb{R}}^{1}$ and height $\ D_h \in {\mathbb{R}}^{1}$ are used. Additionally, $R_A \in {\mathbb{R}}^1$ is used to describe the radius of the multirotor including blades in the XY plane of the frame $\mathscr{F}_B$.

\subsection{Kinematics}
\label{subsec: kinematics}
To derive a unified dynamical model of the UAM and the door while assuring that constraint forces do not appear, Lagrange dynamics are employed. Required kinematic relations for the derivation are listed in this subsection.

With a rotation matrix $R_t$ from the frame $\mathscr{F}_B$ to $\mathscr{F}_W$ and a position vector $\mathbf{d}$ of the end-effector from $O_B$ described in $\mathscr{F}_B$, the kinematic constraint between the UAM and the door can be written as the following:
\begin{equation}
        \mathbf{P} + R_t \mathbf{d} = \mathbf{P_h} + [D_V \cos{\alpha} \ D_V \sin{\alpha} \ D_H]^T
\label{eq: kinematic constraint}
\end{equation}

Defining a new configuration vector $\mathbf{q}=[\mathbf{\Phi} \ \alpha]^T \in {\mathbb{R}}^{4}$, Jacobian matrices mapping $\dot{\mathbf{q}}$ to $\dot{\mathbf{P}}$, $\mathbf{\Omega}$, and $\dot{\alpha}$ can be written as $\dot{\mathbf{P}} = J_t \dot{\mathbf{q}} - R_t \dot{\mathbf{d}}$, ${\mathbf{\mathbf{\Omega}}} = J_r \dot{\mathbf{q}}$, and $\dot{\alpha} = J_{\alpha} \dot{\mathbf{q}}$. Note that $\mathbf{q}$ is chosen to describe the dynamic behavior of the multirotor and the door independent from the motion of the robotic arm, based on the assumption of negligible servomotor dynamics.

\subsection{Dynamics}
\label{subsec: dynamics}
To apply Lagrange equation about $\mathbf{q}$ and $\mathbf{\dot{q}}$, kinetic and potential energies are calculated as follows:
\begin{equation}
\begin{aligned}
    {\mathcal{K}} &=\cfrac{1}{2}\left(\dot{\mathbf{P}}^T M_A \dot{\mathbf{P}} + \mathbf{\Omega}^T I_A \mathbf{\Omega} + \dot{\alpha}^T I_D \dot{\alpha}\right), \\
    {\mathcal{P}} &= m_AgP^Te_3
\end{aligned}
\end{equation}
where $M_A = diag(m_A,m_A,m_A)$, and $e_3 = [0 \ 0 \ 1]^T$. By calculating the Lagrange equation with external forces $\tau_q$, the EoM of the multirotor and the door can be drawn analytically from the below equation:
\begin{equation}
\cfrac{d}{dt}\cfrac{\partial\mathcal{L}}{\partial\dot{q}}-\cfrac{\partial\mathcal{L}}{\partial q} = \tau_q, \ \ \mathcal{L} = \mathcal{K}-\mathcal{P}, 
\label{Lagrange equation}
\end{equation}

The entire system's EoM from the equation (\ref{Lagrange equation}) combined with the motions of the servomotor is as follows:
\begin{equation}
\dot{{\mathbf{x}}}=f({\mathbf{x}},{\mathbf{u}})= 
\begin{bmatrix}
    \dot{\mathbf{q}} \\
    M_{q}^{-1}(-(C_q\dot{\mathbf{q}}+G_q)+\tau) \\
    \dot{\mathbf{H}}_d
\end{bmatrix}
\label{eqn: general EoM}
\end{equation}
where $M_q$, $C_q$, and $G_q$ are mass, Coriolis, and gravitational matrices induced from the equation (\ref{Lagrange equation}), and $\tau = \tau_q + \tau_{ext}$ with $\tau_{ext}$ denoting unmodeled forces and torques from the manipulator and the door. Detailed derivation can be found in \cite{lee2019model} where a sufficiently slow servomotor speed during the entire horizon is assumed. Since servomotors are speed-controlled, with a proper control strategy assuring input limits, the assumption can be satisfied. 

\subsection{Simplified Dynamics}
\label{subsec: simplified dynamics}
A simplified dynamical model for the whole system is derived in this subsection. Necessity for the simplified dynamics can be summarized as follows. Since our approach to handle the door opening scenario with a UAM includes an optimal planner in a receding horizon manner, a more concise model though including all sufficient information of the system dynamics would be more advantageous in terms of computation time. With some additional assumptions, the number of system states can be reduced, and numerical matrix inversion is no longer required which is mandatory in the equation (\ref{eqn: general EoM}).

The simplified version is derived with the following two assumptions:
\begin{itemize}
\item Multirotor's rotational dynamics is negligible.
\item UAM is in a quasi-static state in translational motion during its operation.
\end{itemize}

The first assumption can be justified from the fact that a multirotor's low rotational inertia and its ability to generate high torque ensure that the onboard controller is able to control the angular velocity of the vehicle sufficiently fast. Several other papers \cite{brescianini2018computationally,seo2019robust} include this assumption in their dynamical model for multirotor planning. Secondly, considering the door's relatively higher inertia compared to that of the UAM, the second assumption can be assumed without losing generality. Similar research carried out for operating a drawer with a UAM in \cite{kim2015operating} also assumed this quasi-static motion to generate the desired path. 
Thanks to the first assumption, we can redefine the states and inputs for the system as $\mathbf{x_s} = [\mathbf{\Phi} \ \alpha \ \dot{\alpha} \ \mathbf{H}]^T \in \mathbb{R}^9$, and $\mathbf{u_s} = [f_t \ \mathbf{\Omega}_d \ \mathbf{H}_d]^T \in \mathbb{R}^8$, where $\mathbf{\Omega}_d$ denotes a desired angular velocity of the UAM in $\mathscr{F}_B$. Furthermore, according to the second assumption, introducing an action-reaction force $F_R$ between the UAM and the hinged door, $F_R$ can be written as $F_R = m g e_3 + f_t R_t e_3 $ where $g$ is a gravitational acceleration. Then, the nominal dynamics of the hinged door can be represented as $I_D \Ddot{\alpha}= -{n_D}^T F_R D_V$ where $n_D = [\sin{\alpha} \ -\cos{\alpha} \ 0]^T$ stands for a unit normal vector to the door surface pointing to the opposite direction to the UAM. Therefore, the door's angular acceleration can be rearranged as $\Ddot{\alpha} = g(\mathbf{\Phi},\alpha)f_t$, and the whole system dynamics with the redefined states $\mathbf{x_s}$ and inputs $\mathbf{u_s}$ are as follows:
\begin{equation}
\dot{\mathbf{x}}_s = f_s(\mathbf{x}_s,\mathbf{u}_s) =  
    \begin{bmatrix}
        W(\mathbf{\Phi}) \mathbf{\Omega}_d \\
        \dot{\alpha} \\
        g(\mathbf{\Phi},\alpha) f_t \\
        \dot{\mathbf{H}}_d
    \end{bmatrix}
\label{simplified dynamics}
\end{equation}
where $W(\mathbf{\Phi})$ is a mapping matrix such that $\mathbf{\dot{\Phi}} = W(\mathbf{\Phi})\mathbf{\Omega}$.

\section{TRAJECTORY GENERATION USING \\MODEL PREDICTIVE CONTROL}
\label{sec: trajectory planning}
This section contains ingredients and recipes for a model predictive control problem that produces a desired trajectory to the robust controller. Starting with a problem formulation, hard constraints for avoiding a collision, and an optimal control solver yielding a sub-optimal solution in less than a second, thus enabling a real-time application are presented sequentially. By solving an optimal control problem at every time interval, our MPC algorithm could generate a dynamically feasible and safe trajectory applicable to a real-world experiment.

\subsection{Problem Formulation}
\label{subsec: problem forumlation}
Considering a discretization over time with a time interval $\Delta t$, for the given time horizon $H_O$, an initial state $\mathbf{x}_k$, and an initial input sequence $U_k=\{\mathbf{u}_k, \cdots, \mathbf{u}_{k+H_O-1}\}$ at the $k^{th}$ time element, the MPC problem can be formulated as follows:
\begin{equation}
\begin{aligned}
\min_{U_k} \ & \textrm{J}_k = l_f(\mathbf{x}_{k+H_O}) + \sum_{i=k}^{k+H_O-1}l({\mathbf{x}}_{i},{\mathbf{u}}_{i})\Delta t, \\
\textrm{s.t.} \ & \mathbf{x}_{k+1} = f_D(\mathbf{x}_k,\mathbf{u}_k), \\
              & \mathbf{x}_k \in \mathcal{X}, \\
              & \mathbf{u}_k \in \mathcal{U}
\end{aligned}
\label{problem formulation}
\end{equation}
where $f_D(\mathbf{x}_k,\mathbf{u}_k)$, $\mathcal{X} \subset \mathbb{R}^{n_{\mathbf{x}}}$, and $\mathcal{U} \subset \mathbb{R}^{n_{\mathbf{u}}}$ each denotes a discretized dynamics, a state constraint set, and an input constraint set with $n_{\mathbf{x}} = 9$ and $n_{\mathbf{u}} = 8$. Dynamics $f_D$, states $\mathbf{x}_k$, and inputs $\mathbf{u}_k$ are discretized based on the simplified model in equation (\ref{simplified dynamics}). Since the door's state is included in the system states $\mathbf{x}_k$, through the quadratic cost function with respect to state and input error, state and input trajectories for opening the door while spending minimum energy can be generated. The terminal cost $l_f$ and the stage cost $l$ are defined with each variable's desired value denoted with a superscript of $d$ as follows:
\begin{equation}
\begin{aligned}
l_f(\mathbf{x}_{k+H_O}) =& \cfrac{1}{2} \|\mathbf{x}_{k+H_O}-\mathbf{x}^d_{k+H_O} \|^2_L, \\
l(\mathbf{x}_i,\mathbf{u}_i) =& \cfrac{1}{2} \|\mathbf{x}_{i}-\mathbf{x}^d_{i} \|^2_Q + \cfrac{1}{2} \|\mathbf{u}_{i}-\mathbf{u}^d_{i} \|^2_R
\end{aligned}
\label{cost weight}
\end{equation}

\subsection{Hard Constraints}
\label{subsec: hard constraints}
To ensure a safe operation for the scenario, several hard constraints on states should be considered. In this section, three types of constraints for collision avoidance are handled: self-collision avoidance, door collision avoidance, and doorframe collision avoidance. These constraints are first derived using kinematic relationship and organized into a state constraint set $\mathcal{X}$ as in equation (\ref{problem formulation}). No hard constraints on inputs are considered in this paper; therefore, the input constraint set is set to be $\mathcal{U} = \mathbb{R}^{n_{\mathbf{u}}}$.                                                            
\begin{figure}
    \centering
    \includegraphics[width=0.6\linewidth]{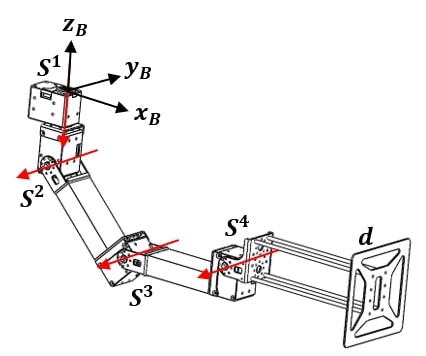}
    \caption{Configuration of the robotic arm.}
    \label{fig:robotic_arm}
    \vspace{-0.3cm}
\end{figure}
First of all, unlike a general multirotor, our UAM's additional freedom in the robotic arm should be carefully managed to avoid a crash between the multirotor airframe and the robotic arm. To ensure this self-collision avoidance, following constraints are devised:
\begin{equation}
\begin{aligned}
    S^3_z \leq 0 \\
    S^4_z \leq 0 \\
    d_z \leq 0 
\end{aligned}
\label{constraint_1}
\end{equation}
where $S^3$, $S^4$, and $d$ are position vectors of the $3^{rd}$ servomotor, the $4^{th}$ servomotor, and the end-effector of the robotic arm described in $\mathscr{F}_B$ while having their origins at $O_B$ as in the Fig. \ref{fig:robotic_arm}, and the subscript $z$ in $*_z$ denotes the third component of a vector $*$. Since $O_B$ is assumed to be centered at the CoM of the multirotor, and these vectors are all described in $\mathscr{F}_B$, the above constraints imply that the robotic arm must always stay below the airframe of the multirotor. Note that all three position vectors $S^3$, $S^4$, and $d$ are a function of $\mathbf{H}$ which can be derived with forward kinematics; therefore, these constraints can be formulated only with system states. 

The second constraint, which is avoiding collision with the door, is constructed as follows:
\begin{equation}
\begin{aligned}
    n_D^T(R_t d) \geq R_A \max_{\theta} \left\{n_D^T\left(R_t \begin{bmatrix}
                                                    \cos\theta \ \sin\theta \ 0
                                                    \end{bmatrix}^T\right)\right\}
\end{aligned}
\label{constraint_2}
\end{equation}
In equation (\ref{constraint_2}), the left-hand side indicates the shortest distance between the CoM of the UAM and the door surface, and the right-hand side quantifies the distance between the CoM of the UAM and the multirotor's airframe closest to the door surface. If we introduce $n^B_D = R_t^T n_D \in \mathbb{R}^3$, a door surface unit normal vector in $\mathscr{F}_B$, the equation (\ref{constraint_2}) can be arranged as follows:
\begin{equation}
\begin{aligned}
    (n^B_D)^T d \geq R_A \sqrt{(n^B_{Dx})^2 + (n^B_{Dy})^2}
\end{aligned}
\label{constraint_2_simplified}
\end{equation}
Since all variables in this constraint are functions of states $x_s$ only, this second constraint can also be formulated only with the system states.

The last doorframe avoiding constraint is formulated as the following:
\begin{equation}
\begin{aligned}
    P_{Hy} + R_A \leq P_y \leq P_{Hy} + D_w - R_A
\end{aligned}
\label{constraint_3}
\end{equation}
where subscript $y$ of $*_y$ denotes the second component of a vector $*$.
This equation can be further organized and can only be expressed with states as 
\begin{equation}
\begin{aligned}
    D_V \sin\alpha +R_A + D_w \leq d^W_{y} \leq D_V \sin\alpha - R_A
\end{aligned}
\label{constraint_3_simplifed}
\end{equation}
where $d^W(\mathbf{\Phi},\mathbf{H}) = R_t d$ is the end-effector position vector from the CoM of the UAM described in the frame $\mathscr{F}_W$.

\subsection{Optimal Control Solver}
\label{subsec: optimal control solver}
Based on the problem and constraints formulated in the subsection \ref{subsec: problem forumlation} and \ref{subsec: hard constraints}, the existing algorithm of the constrained version of differential dynamic programming in \cite{plancher2017constrained} is employed to handle nonlinear dynamics with state constraints. This solver could generate an optimized nominal trajectory satisfying constraints in about 30 Hz with a time horizon of 1 second. 

\section{EXPERIMENTS}
\label{sec: experiments}
\subsection{Control Framework}
\label{subsec: control framework}
\begin{figure}
    \centering
    \vspace{0.3cm}
    \includegraphics[width=1.0\linewidth]{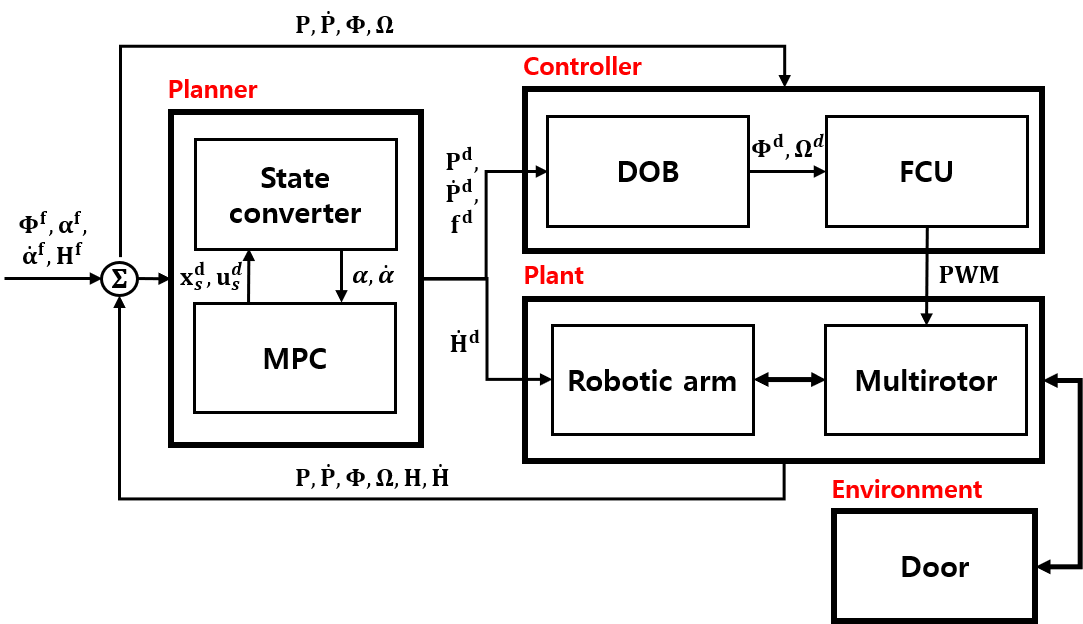}
    \caption{Overall control and planning structure of the aerial manipulation system.}
    \label{fig: control framework}
\end{figure}
Overall control framework is illustrated in the Fig. \ref{fig: control framework}. $\mathbf{\Phi}$, $\alpha$, $\dot{\alpha}$, and $\mathbf{H}$ with $f$ as their superscript denotes final target states $\mathbf{x}^f_s$. Since only observable states from the UAM are $\mathbf{P}$, $\dot{\mathbf{P}}$, $\mathbf{\Phi}$, $\mathbf{\Omega}$, $\mathbf{H}$, and $\mathbf{\dot{H}}$, states for the door have to be converted in state converter in Fig. \ref{fig: control framework} as:
\begin{equation}
\begin{aligned}
    \alpha &= \arctan\left(\cfrac{-\mathbf{P}_{hy}+\mathbf{P}_y + d^W_y}{-\mathbf{P}_{hx}+\mathbf{P}_x + d^W_x}\right) \\
    \dot{\alpha} &= 
    \begin{cases}
        \cfrac{\dot{\mathbf{P}}_x + d_{Rx}^W + \dot{d}^W_x}{-D_V\sin{\alpha}}, & \text{if $\alpha \neq n\pi$} \\
        \cfrac{\dot{\mathbf{P}}_y + d_{Ry}^W + \dot{d}^W_y}{D_V\cos{\alpha}}, & \text{otherwise}
    \end{cases}
\end{aligned}
\label{eqn: state conversion}
\end{equation}
where $n \in \mathbb{Z}$, $\dot{d}^W = R_t \dot{d}$, and $d_R^W = R_t \hat {\Omega}d$. $\hat{*}$ is a hat operator describing a cross product between two vectors as $\hat{a}b = a \times b$. This state conversion is based on the kinematic equation (\ref{eq: kinematic constraint}) and its time derivatives.

After the MPC module receives all state information, it generates state and input trajectories for certain time steps $H \Delta t$. These trajectories are translated back into the aerial manipulation's desired position and velocity by the state converter, using a similar process as in the equation (\ref{eqn: state conversion}). Among the series of desired position and velocity for both the multirotor and the robotic arm's servomotors, the first arrived desired setpoints are subscribed to the controller as new setpoints. The robotic arm in the flow chart contains an inherent velocity controller, and therefore desired velocity is depicted to be directly published to the robotic arm. 

Stability is one of the inevitable and challenging problems in controlling an aerial manipulator. Although the input trajectory subscribed from MPC could be directly applied to the flight controller (FCU) for attitude control, which is Pixhawk 2 in our case, it is hard to guarantee stability during flight. Consequently, we adopt a disturbance observer-based robust controller implemented in \cite{kim2017robust} for position control where its stability is fully analyzed along with experimental validations. This controller generates the desired attitude and angular velocities again to the flight controller by which we can ensure the aerial manipulation's stability.


\subsection{Experimental Setup}
\label{subsec: experimental setup}
We use a DJI F550 multirotor frame with 2312E motors controlled by 420Lite electronic speed controllers. The robotic arm is composed of ROBOTIS dynamixel XH430 series, and frames of OPEN MANIPULATOR-X. As the onboard computer, Intel NUC running Robot Operating System (ROS) in Ubuntu 16.04 executes MPC-based trajectory planner, DOB-based robust position controller, and navigation algorithm with VICON. As in Fig. \ref{fig: control framework}, the vehicle's attitude is controlled by the onboard low-level controller in Pixhawk 2. In addition, to address a realistic scenario, a door with its width and height of $D_w = 1.2 \ m$, and $D_h = 1.6 \ m$ is employed. The weight of the door is about $11 kg$, and with these values, the mass moment of inertia of the door can be calculated as $I_D \approx 5.28 \ kgm^2$. 

Weight matrices, desired final states and inputs for the optimal problem are designed as 
\begin{equation}
\begin{aligned}
    L = Q =& diag[5 \ 5 \ 3 \ 9 \ 8 \ 0.05 \ 0.1 \ 0.1 \ 0.1], \\
    R =& diag[0.1 \ 5 \ 5 \ 13.5 \ 10 \ 10 \ 10 \ 10], \\
    \mathbf{x}_f =& [0 \ 0 \ \cfrac{-7\pi}{18} \ \cfrac{\pi}{9}  \ 0 \ 0 \ \cfrac{\pi}{2} \ -\cfrac{\pi}{2} \ 0]^T
\end{aligned}
\end{equation}
where $\mathbf{x}_f$ denotes a final desired state which is used as $\mathbf{x}^d_{k+H_O} = \mathbf{x_f}$. We set the initial door state as $\alpha = \pi/2$ and $\dot{\alpha} = 0$. 

\begin{figure}
    \centering
    \includegraphics[width=0.95\linewidth]{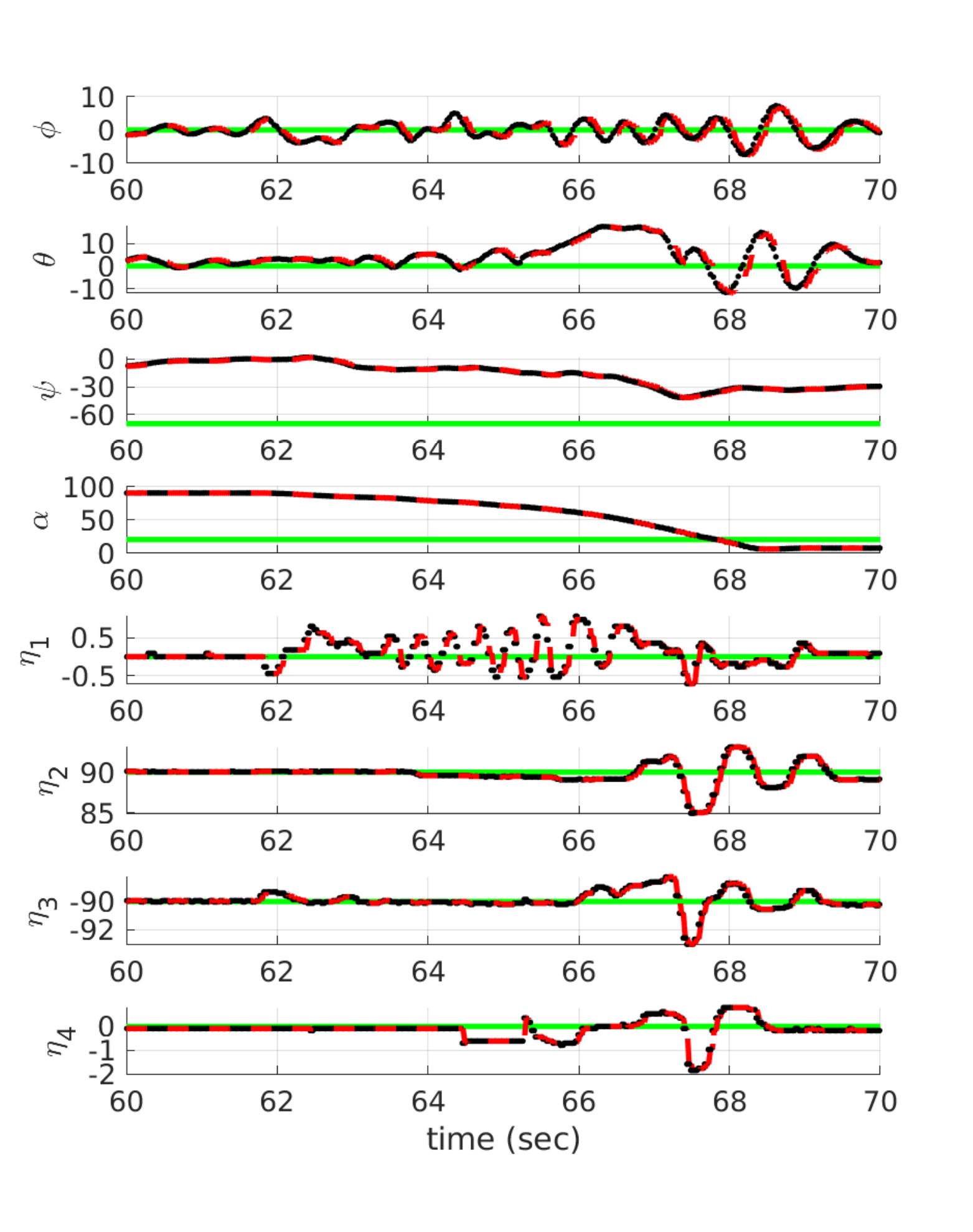}
    \vspace{-0.8cm}
    \caption{History of the states during the door opening experiment. The black line represents measured value. The dashed red line describes the predicted state from the MPC module. The green line represents the desired state of the system $\mathbf{x}_f$. All units on the y axis are degree.}
    \label{fig: results}
    \vspace{-0.3cm}
\end{figure}

\subsection{Results}
\label{subsec: results}
In the experiment, an aerial manipulator pushing a customized hinged door is conducted. 
Thanks to the capability of considering the state constraints in planning trajectory, the aerial manipulator successfully opens the door without collision as illustrated in Fig. \ref{fig:thumbnail}. 
The history of the states during the experiment is described in Fig. \ref{fig: results}.
As expected, the door angle $\alpha$ tends to converge to the desired final value, implying that the door is sufficiently opened. 
Furthermore, followed by the changes in the door angle $\alpha$, the vehicle's yaw motion $\psi$ rotates accordingly.
However, discrepancies between the desired and measured states, especially in $\alpha$ and $\psi$, occur due to the fact that the assumption 3 happens to be violated intermittently during the experiment. Although it is assumed that the end-effector is firmly attached to the door surface, uncertainty in door parameters and unmodeled dynamics between the UAM and the door seem to cause a faster door movement which results in a detachment between the door surface and the end-effector. Force control strategy like impedance control seems to be capable of handling this problem, and we leave it as a future work.


\section{CONCLUSION}
\label{sec: conclusion}
In this paper, a systematic methodology for an aerial manipulator operating a hinged door is presented. Coupled equations of motion encompassing the aerial manipulator and the hinged door are first derived and later simplified to be applicable to an online-solvable optimal control problem. State constraints guaranteeing a safe trajectory are proposed, and the formulated MPC problem is solved with a constrained DDP algorithm. Generated trajectory is then tracked by a DOB-based robust controller, which provides stability during execution. For future studies, along with the one mentioned in the subsection \ref{subsec: results}, a robotic arm with higher degrees of freedom in the end-effector is anticipated to provide better maneuverability while in interaction.

\end{document}